\documentclass[sigconf]{acmart}
\usepackage[utf8]{inputenc}
\usepackage[super]{nth}
\usepackage{amssymb,amsmath,amsfonts,amsthm,enumitem,bm}
\usepackage{verbatim}
\usepackage{url}

\expandafter\def\expandafter\normalsize\expandafter{%
    \normalsize
    \setlength\abovedisplayskip{0pt}
    \setlength\belowdisplayskip{1pt}
    \setlength\abovedisplayshortskip{-5pt}
    \setlength\belowdisplayshortskip{4pt}
}

\begin{document}

\title{Long Distance Relationships without Time Travel: Boosting the Performance of a Sparse Predictive Autoencoder in Sequence Modeling}

\author{Jeremy Gordon}
\thanks{Work completed while the first author was completing a summer placement with Numenta}
\affiliation{UC Berkeley \& Numenta}
\email{jrgordon@berkeley.edu}

\author{David Rawlinson}
\affiliation{Incubator 491}
\email{dave@agi.io}

\author{Subutai Ahmad}
\affiliation{Numenta}
\email{sahmad@numenta.com}

\date{October 2019}

\begin{abstract}
In sequence learning tasks such as language modelling, Recurrent Neural Networks must learn relationships between input features separated by time. State of the art models such as LSTM and Transformer are trained by backpropagation of losses into prior hidden states and inputs held in memory. This allows gradients to flow from present to past and effectively learn with perfect hindsight, but at a significant memory cost. In this paper we show that it is possible to train high performance recurrent networks using information that is local in time, and thereby achieve a significantly reduced memory footprint. We describe a predictive autoencoder called bRSM featuring recurrent connections, sparse activations, and a boosting rule for improved cell utilization. The architecture demonstrates near optimal performance on a non-deterministic (stochastic) partially-observable sequence learning task consisting of high-Markov-order sequences of MNIST digits. We find that this model learns these sequences faster and more completely than an LSTM, and offer several possible explanations why the LSTM architecture might struggle with the partially observable sequence structure in this task. We also apply our model to a next word prediction task on the Penn Treebank (PTB) dataset. We show that a `flattened’ RSM network, when paired with a modern semantic word embedding and the addition of boosting, achieves 103.5 PPL (a 20-point improvement over the best N-gram models), beating ordinary RNNs trained with BPTT and approaching the scores of early LSTM implementations. This work provides encouraging evidence that strong results on challenging tasks such as language modelling may be possible using less memory intensive, biologically-plausible training regimes.

\end{abstract}

\maketitle

\section{Introduction}

In the sequence learning domain, the challenge of modeling relationships between related elements separated by long temporal distances is well known. Language modeling, the task of next character or next word prediction, is an extensively studied paradigm that exhibits the need to capture such long-distance relationships that are inherent to natural language. Historically, a variety of architectures have achieved excellent language modelling performance. Although larger datasets and increased memory capacity have also improved results, architectural changes have been associated with more significant improvements on older benchmarks.

N-gram models are an intuitive baseline model and were developed early in this history. N-gram models learn a distribution over the corpus vocabulary conditioned on the $n$ prior tokens, e.g. a tri-gram ($n=3$) model makes predictions based on the distribution: 

$$ P(x_i|x_{t-3}, x_{t-2}, x_{t-1}) = f(x_{t-3}, x_{t-2}, x_{t-1}) $$

Among N-gram models, smoothed 5-gram models achieve minimum perplexity on the Penn Treebank dataset \cite{marcus1994penn}, a result that illustrates constraints on the value of increasingly long temporal context.

More recent approaches have demonstrated the success of neural models such as Recurrent Neural Networks applied to language modeling. In 2011, Mikolov et al. presented a review of language models on the Penn Tree-bank (PTB) corpus showing that recurrent neural models at that time outperformed all other architectures \cite{mikolovExtensionsRecurrentNeural2011}. 

Ordinary RNNs are known to suffer from the vanishing gradient problem in which partial derivatives used to backpropagate error signals across many layers approach zero. Hochreiter et al introduced a novel multi-gate architecture called Long Short-Term Memory (LSTM) as a potential solution \cite{hochreiterLongShortTermMemory1997}. Models featuring LSTM have demonstrated state of the art results in language modeling, demonstrating their ability to robustly learn long-range causal structure in sequential input.

Though RNNs appear to be a natural fit for language modeling due to the inherently sequential nature of the task, feed-forward networks utilizing novel convolutional strategies have also been competitive in recent years. WaveNet is a deep autoregressive model using dilated causal convolutions in order to achieve long temporal range receptive fields \cite{oordWaveNetGenerativeModel2016}. A recent review compared the wider family of temporal convolutional networks (TCN)---of which WaveNet is a member---with recurrent architectures such as LSTM and GRU, finding that TCNs surpassed traditional recurrent models on a wide range of sequence learning tasks \cite{bai2018empirical}.

Extending the concept of replacing recurrence with autoregressive convolution, Vaswani et al. added attentional filtering to their Transformer network \cite{vaswaniAttentionAllYou2017}. The Transformer uses a deep encoder and decoder each composed of multi-headed attention and feed-forward layers. While the dilated convolutions of WaveNet allow it to learn relationships across longer temporal windows, attention allows the network to learn which parts of the input, as well as intermediate hidden states, are most useful for the present output.

Current state-of-the-art results are achieved by GPT-2, a 1.5 billion parameter Transformer \cite{gong2018frage}, which obtains 35.7 PPL on the PTB task (see Table \ref{tab:lm_results}). The previous state of the art was an LSTM with the addition of mutual gating of the current input and the previous output reporting 44.8 PPL \cite{melisMogrifierLSTM2019}.

Common to all the neural approaches reviewed here is the use of some form of deep-backpropagation, either by unrolling through time (see section \ref{section:tbptt} for more detail) or through a finite window of recent inputs (WaveNet, Transformer). Since most of these models also benefit from deep multilayer architectures, backpropagation must flow across layers, and over time steps or input positions, resulting in very large computational graphs across which gradients much flow. By contrast, all other methods in the literature (such as traditional feed-forward ANNs and N-gram models) are not known to produce such good performance (i.e. none have surpassed 100 PPL on PTB). 

\subsection{Motivation}

Despite the impressive successes of the recurrent, autoregressive, and attention-based approaches reviewed above, the question remains whether similar performance can be achieved by models that do not depend on deep backpropagation. Models that avoid backpropagation across many layers or time steps are interesting for two reasons. First, computational efficiency is becoming an increasingly important consideration in deep learning, both due to the pragmatics of designing algorithms that must be trained in resource constrained environments such as edge computing, and as researchers begin to acknowledge the significant environmental footprint of the hardware that drives machine learning at scale \cite{haoTrainingSingleAI}. Second, to the extent that computational models may help us better understand the dynamics and perhaps mechanisms underlying our own cognitive abilities, architectures constrained by similar principles as those that govern the brain may offer more credible insights. Specifically, we are interested in models that lie within the biologically plausible criteria outlined by Rawlinson et al.: 1) local and immediate credit assignment, 2) no synaptic memory, and 3) no time-traveling synapses \cite{rawlinsonLearningDistantCause2019}. Our goal, then, is to explore and push the performance bounds of sequence learning models leveraging dynamics consistent with these bio-plausibility constraints.

\section{Method}

\subsection{Original RSM Model}
\label{sec:orsm}

\begin{figure}[h]
    \centering
    \includegraphics[width=\linewidth]{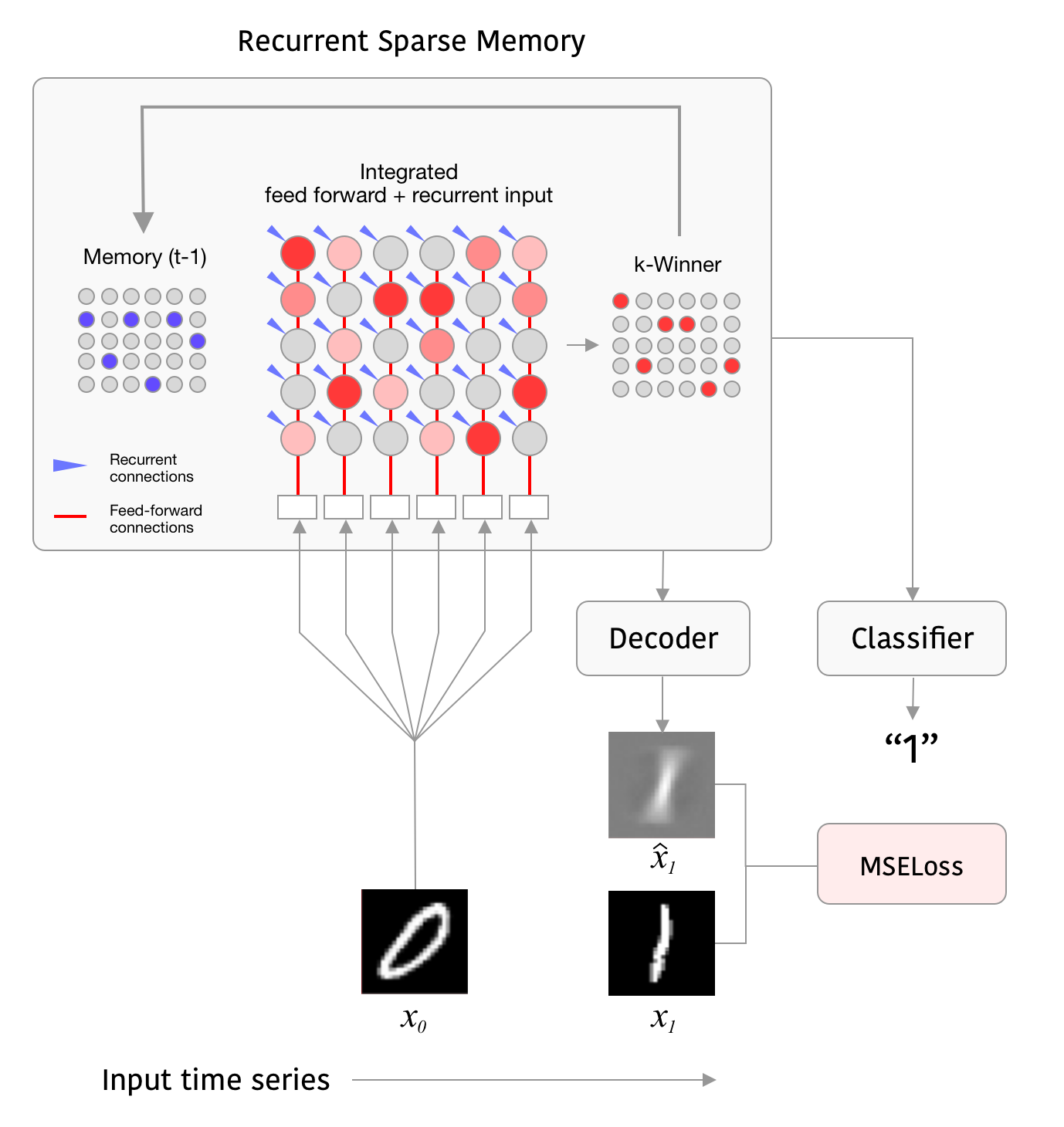}
    \caption{Schematic of original RSM architecture, shown processing inputs from the stochastic sequential MNIST task (see section \ref{section:exp_ssmnist}).  Note that, as per original paper, the RSM network is trained only on the MSE loss, and is not affected by gradients backpropagated from the classifier network.}
    \Description{Schematic of original RSM architecture. Note that, as per original paper, no gradients pass from the classifier to the RSM in order to keep credit assignment local.}
    \label{fig:rsm}
\end{figure}

We began with the Recurrent Sparse Memory (RSM) architecture proposed by Rawlinson et al. \cite{rawlinsonLearningDistantCause2019}. RSM is a predictive recurrent autoencoder that receives sequential inputs (e.g. images or word tokens), and is trained to generate a prediction of the next item in the sequence (see schematic in Figure \ref{fig:rsm}). Like Hierarchical Temporal Memory \cite{Hawkins2016}, the RSM memory is organized into $m$ groups (or mini-columns), each composed of $n$ cells. Cells within each group share a single set of weights from feed forward input, such that the feed-forward contribution $\bm{z}^A$ is an $m$-dimensional vector computed as:

$$\bm{z}^A = w^A\bm{x}^A(t)$$

Each cell receives dense recurrent connections from all cells at the previous time step, and the recurrent contribution $z^B$ is an $m \times n$ matrix computed as:

$$z^B = w^B\bm{x}^B(t)$$

$\sigma_{ij}$ is an $m \times n$ matrix holding the weighted sum combining feed-forward and recurrent input to each cell $j$ in group $i$, and is given by:

$$\sigma_{ij} = \bm{z}^A_i + z^B_{ij}$$

A top-\emph{k} sparsity is used as per Makzhani and Frey \cite{makhzaniKSparseAutoencoders2013}. RSM implements this sparsity by computing two sparse binary masks, $M^\pi$ and $M^\lambda$, which indicate the most active cell (one per group), and most active group ($k$ per layer), respectively. An inhibition trace was used in the original model to encourage efficient resource utilization during the sparsening step, but is replaced with boosting in this work (see section \ref{sec:resource_utilization} for discussion). The final output is calculated by applying a $tanh$ nonlinearity to the sparsened activity:

$$\bm{y}_{ij} = \tanh( \sigma_{ij} \cdot M_{i}^\lambda \cdot M_{ij}^\pi )$$

A memory trace $\bm{\psi}(t)$ is maintained with an exponential decay parameterized by $\epsilon$, such that $\bm{\psi}(t) = \max( \bm{\psi}(t-1) \cdot \epsilon, \bm{y} )$. From $\bm{\psi}$, the recurrent input at the next time step is calculated by normalizing with constant $\alpha$, chosen such that the activity in $\bm{x}^B$ sums to 1: 

$$\bm{x}^B(t+1) = \alpha \cdot \bm{\psi}(t)$$

Like other predictive autoencoders, RSM is trained to generate the next input $\bm{\hat{x}}^A$ by ``decoding'' from the max of each group's sparse activity:

$$\bm{y}^\lambda_{i} = \max( \bm{y}_{i1}, \dotso, \bm{y}_{in} )$$

The prediction is then computed as $\bm{\hat{x}}^A(t) = w^D\bm{y}^\lambda$, where $w^D$ is a weight matrix with dimension equal to the transpose of $w^A$. 

Finally, to read out labels or word distributions from the network, RSM uses a simple classifier network composed of a 2-layer fully connected ANN using leaky ReLU nonlinearities. The classifier network is trained concurrently but independently to the RSM network (not sharing gradients), and takes the RSM's hidden state as input.

\subsection{Boosted RSM (bRSM)}
\label{sec:additions}

We developed a variant of RSM that (among other architectural changes) replaces cell-inhibition with a cell activity `boosting' scheme. For brevity, we refer to the modified algorithm as bRSM.

In an attempt to encourage better generalization, we explored a number of adjustments to the original model described in section \ref{sec:orsm}. Additional model details and hyper-parameter settings for reported experiments are included in Appendix \ref{appendix:model_details}, and the full code for all experiments is publicly accessible\footnote{The full code for the bRSM model and all experiments is available at 
\url{https://github.com/numenta/nupic.research/tree/master/projects/rsm}}.

We find that bRSM significantly improves performance on the language modeling task. We review each of our adjustments in the section below.

\subsubsection{Flattened network}

A fundamental dynamic of HTM-like architectures is that each mini-column learns some spatial structure in the input, and each cell within a mini-column learns a transition from a prior representation \cite{Hawkins-et-al-2016-Book}. A potential limitation of this architecture is that, while representations of the input via feed forward connections benefit from spatial semantics (similar representations for similar inputs), the predictive representations developed through recurrent connections lack this property: similar sequence items in different sequential contexts are highly orthogonal \cite{rawlinsonLearningDistantCause2019}.

To illustrate a potential inefficiency of this orthogonality, consider a network trained on sequences where some set of similar inputs $A = \{A_1, A_2, A_3\}$ predict both $B$ and $C$ at the next time step, prompting cells in the representations of both $B$ and $C$ to activate when exposed to inputs in $A$. These cells may contain nearly identical weights linked to a sparse representation generalizing across patterns in $A$. Such a redundancy might be avoided if some subset of cells having learned the transition from $A$ could be shared by both $B$ and $C$. This line of reasoning motivated experiments in which each group was set to have only one cell, thus removing shared feed-forward weights from the model, and enabling decoding from the full hidden state rather than a group-max bottleneck. The flexibility of allowing predictive cells to participate in multiple input representations may explain the improved performance of this flattened architecture in the language modeling task, though we suspect the grouped model may be beneficial on tasks with higher-order compositionality in space or time.

\subsubsection{Boosting}
\label{sec:boosting}

Sparse networks may learn locally optimal configurations in which only a small fraction of a layer's representational capacity is used. When this occurs, many units remain idle resulting in inefficient resource usage and limited performance. The original RSM model employs an inhibition strategy whereby a separate exponentially decaying trace is used to discourage recently active cells from re-activating.

An alternative strategy known as boosting has been proposed to achieve the same goal but exhibits different properties from inhibition. We used a boosted k-Winners algorithm suggested by Cui et al. \cite{ahmadHowCanWe}. This algorithm tracks the duty cycle of each cell $d_i$, which captures the probability of recent activation (sparsened via top-\emph{k} masking):

$$ d_i(t) = (1-\alpha) \cdot d_i (t-1) + \alpha [i \in topIndices] $$

A per-cell boost term $b_i$ is then computed based on this duty cycle, increasing the probability of less recently active cells from firing, and inhibiting those more recently active:

$$ b_i(t) = e^{\beta (\hat{a}-d_i(t))} $$ where $\hat{a}$ is the expected layer sparseness defined as the number of winners divided by the layer size, $\frac{k}{mn}$, and $\beta$ is the boost strength hyper-parameter which can be optionally configured to remain fixed or decay during training (see Appendix \ref{appendix:model_details}). The per-cell weighted sum $\sigma_{ij}$ is then redefined as:

$$\sigma_{ij} = ( \bm{z}^A_i + z^B_{ij} ) \cdot b_i$$

\subsubsection{Semantic embedding}

Rawlinson et al. tested RSM with a synthetic binary word embedding (see Appendix \ref{appendix:synthetic_embedding}) with no semantic properties in order to isolate the performance of the architecture from that of the embedding. Since RSM was not specifically designed to learn high quality language embeddings, we chose to use a modern embedding leveraging sub-word semantics. We pretrained a 100-dimensional FastText \cite{bojanowski2016enriching} embedding on the training corpus, and used this as input for all experiments (see Appendix \ref{appendix:fasttext} for generation details).  

\subsubsection{Trainable decay}

In language modeling, some tokens may provide useful context to word prediction many tokens in the future (e.g. rare words unique to a particular topic), while others may be necessary for next word prediction (e.g. tokens composing multi-word proper nouns or phrases, or common words indicating syntactic structure). In the original RSM model, the rate of decay of the recurrent input is parameterized by a single scalar value $\epsilon$, which is multiplied into the prior memory state on each time step. While each cell participates in multiple input representations, it may be possible to improve generalization performance by learning a unique exponential decay scalar for each cell in the memory. We implemented trainable decay as a single tensor $\Delta$ of dimension $m * n$ (equivalent to just $m$ in the flattened architecture), which we pass through a Sigmoid before applying to the memory in the decay step:

$$ \bm{\psi}(t+1) = \bm{\psi}(t) \cdot \sigma(\Delta) $$

We found that applying a ceiling close to 1 to the $\sigma(\Delta)$ term helped to avoid volatility likely caused by the memory state retaining too much history.

The benefit of moving to a trainable decay parameter requires a nominal increase in parameters, and provides a consistent but small improvement (\textasciitilde 5 PPL on next word prediction).

\subsubsection{Functional Partitioning}

We found one final addition to be significantly beneficial on the stochastic sequential MNIST task (detailed in section \ref{section:exp_ssmnist}). In this version of the model, the bRSM memory is partitioned into either two or three blocks: one taking feed-forward input only, one taking recurrent input only, and one integrating both input sources via addition. This third section is equivalent to the full memory in the original RSM model. To ensure utilization across all partitions while keeping target sparsity consistent, we applied the top-k nonlinearity to each partition separately, with partition winners $k_p$ proportional to partition cell count $m_p$:

$$k_p = k \frac{m_p}{m}$$ 

The motivation behind functional partitioning was an extension of the logic behind the use of a flattened memory. To the extent that it is useful for some cells to represent transitions from prior input, and others to represent current input, we wondered if an architecture in which these functional roles are enforced would improve performance. 

The partitioned model whose ssMNIST results appear in Figure \ref{fig:ssmnist_acc} uses a memory with cells allocated as follows: 7\% feed-forward, 85\% recurrent, and 8\% integrated. The resultant model contains fewer parameters since a portion of cells are connected only to the input, which has lower dimensionality than the full memory.

This partitioning method did not improve generalization on the language modeling task hence these results are not reported.

\section{Experiments}

\subsection{Tasks \& Datasets}
We selected tasks anticipated to be difficult for RNNs and RSM in particular, to enable empirical characterization of its limitations. We tested bRSM on two tasks: a non-deterministic version of the original partially-observable MNIST sequence task \cite{rawlinsonLearningDistantCause2019}, as well as next word prediction (language modeling) on the Penn Treebank dataset.

\subsubsection{Stochastic Sequential MNIST (ssMNIST)}
\label{section:exp_ssmnist}

RSM was initially tested on a partially observable sequence learning task in which the network is exposed to higher-order sequences of randomly chosen MNIST images drawn according to a \emph{predetermined} list of labels e.g. ``0123 0123 0321''. It is then possible that algorithms could learn to ignore the images and simply keep count to make accurate predictions. A potential expansion to this task, then, is to require memorization of repeatable sub-sequences (e.g. the 12 digit example above) presented in a \emph{random} order. This requires repeatable sub-sequences to be learned, while also learning to ignore sub-sequence order that has no predictive value. The image-observations and transitions are then both partially non-deterministic, and the images must be considered for optimal accuracy.

These randomly ordered sub-sequences can be described by a grammar. The grammar generating process is configured to specify $m$ sub-sequences of length $n$ digits each. Details of the grammar generated are described in Appendix \ref{appendix:sequences}.

Using a single fixed grammar we can construct an observation generating process that randomly chooses between sub-sequences, but then follows each sub-sequence deterministically, as follows:

\begin{enumerate}
    \item Select one sub-sequence from the $m$ specified uniformly at random
    \item Select the first digit label in the sub-sequence
    \item Select a random MNIST digit according to the selected label
    \item Move through the sub-sequence, drawing random MNIST digits for each label, until the end is reached
    \item Go to 1
\end{enumerate}

\begin{figure}[h]
    \centering
    \includegraphics[width=\linewidth]{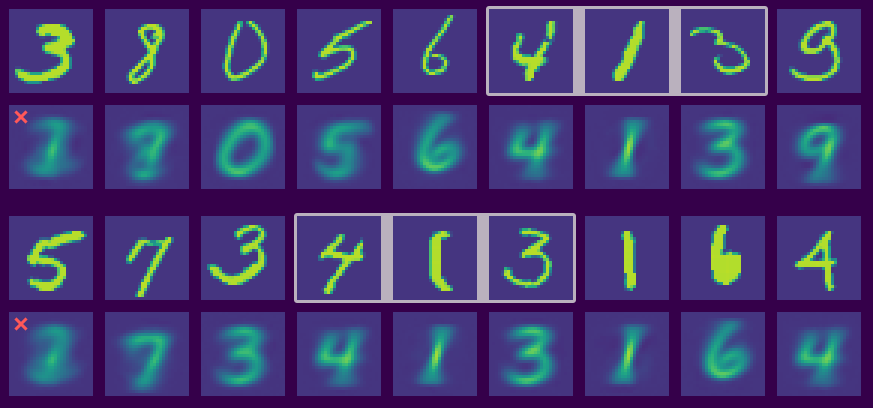}
    \caption{High-order, partially observable stochastic sequence learning predictions. Rows alternate between actual 9-digit samples from the grammar, and bRSM predictions. Sequences ``6-4-1-3-9" and ``3-4-1-3-1" (with common sub-sequence ``4-1-3" outlined) are predicted correctly.}
    \label{fig:ssmnist_sequence}
\end{figure}

We generated a test grammar composed of 8 sub-sequences of 9 MNIST digits each (dimension specified to minimize confusion, see sample sequence and predicted outputs in Figure \ref{fig:ssmnist_sequence}). This specific ``8x9'' grammar for which we report results, along with a calculation for the theoretical limit on prediction accuracy, is included in Appendix \ref{appendix:sequences}.

To ensure that solving the task would require the successful learning of higher order sequences, we confirmed that prediction of at least some of the transitions in the resultant grammar required knowledge of the sequence item two or more steps prior.

Unlike many RNN tasks, there is no flag or special token to indicate sub-sequence boundaries or task reset. Without any priors for the length or existence of sub-sequences, the ssMNIST task is challenging even for humans.

\subsubsection{Baseline: tBPTT trained LSTM}
\label{section:tbptt}

We chose to use an LSTM as a `baseline' algorithm to represent the deep-backpropagation approach and compare to bRSM. Modern recurrent neural networks such as LSTMs are trained using backpropagation through time (BPTT), which conceptually unrolls the network's computational graph across multiple time steps resulting in a standard multi-layer feed-forward network, and then backpropagating the loss from one or more output layers (or heads) towards the shallower layers representing earlier timesteps.

The LSTM was trained with Adam using a learning rate of $2 \times 10^{-5}$. We set the hidden size of the LSTM layer to produce networks roughly consistent with the parameter count of bRSM. Results reported below are for an LSTM with 450 hidden units (2.57M parameters).

\begin{figure}[h]
    \centering
    \includegraphics[width=\linewidth]{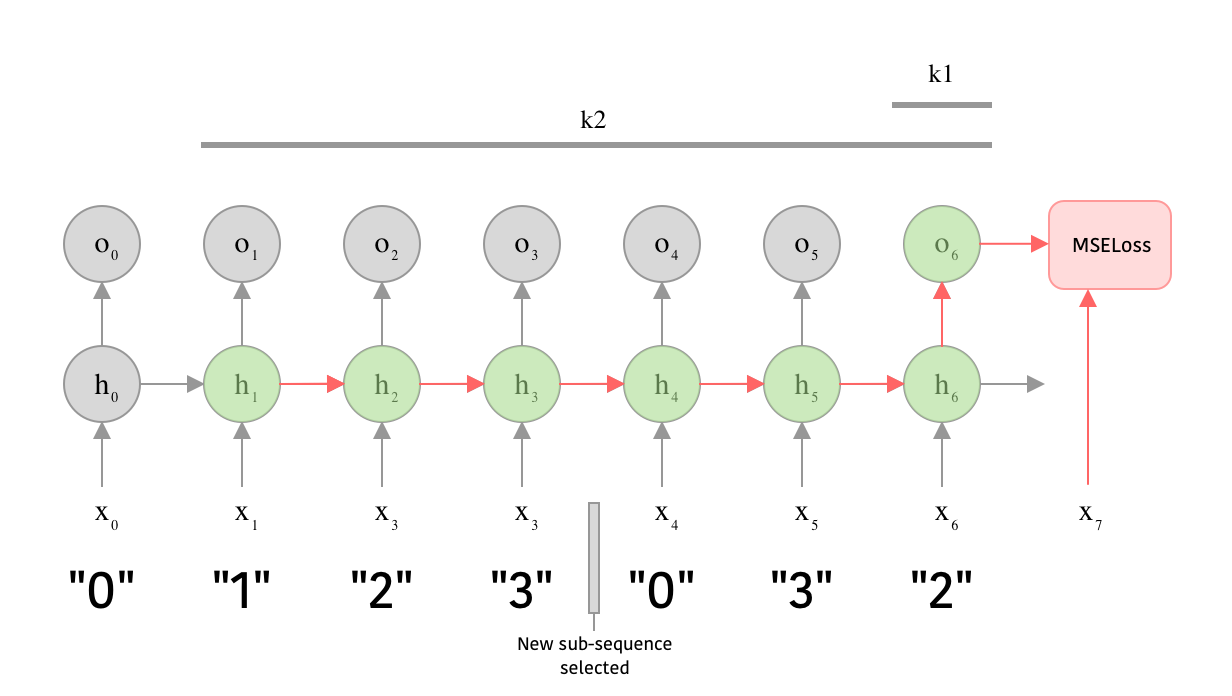}
    \caption{Schematic of truncated backpropagation through time parameterization BPTT($k_1$, $k_2$), with $k_1$=1, $k_2$=6 for simple grammar [\{0123\}, \{0321\}]. $x$, $h$ and $o$ represent input, hidden and output respectively.}
    \label{fig:bptt}
\end{figure}

We implemented a training regime consistent with Williams and Peng's improved truncated-BPTT algorithm  \cite{williamsEfficientGradientBasedAlgorithm} which is parameterized by two integers determining the flow of gradients through past states of the network. In $tBPTT(k_{1}, k_{2})$, $k_{1}$ specifies the interval at which to inject error from the last $k_{1}$ outputs, while $k_{2}$ specifies the length of the history through which gradients should propagate. We set $k_{1}=1$ to match the online ``one digit, one prediction'' dynamic of the ssMNIST task. After disappointing initial results with large $k_{2}$ values, we experimented with a range of values to empirically optimize LSTM performance (see schematic in Figure \ref{fig:bptt}). 

\begin{figure}[h]
    \centering
    \includegraphics[width=\linewidth]{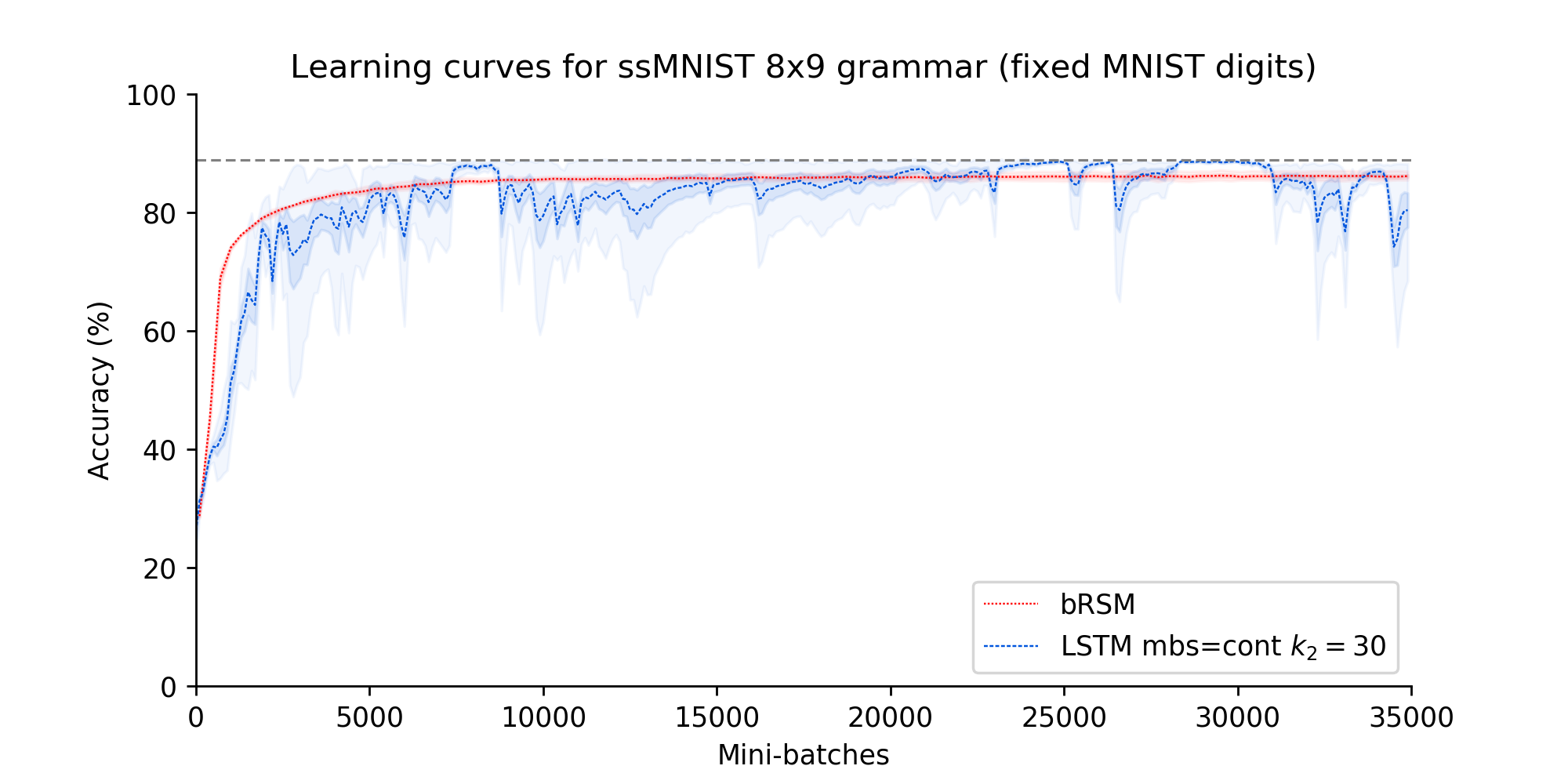}
    \caption{LSTM and bRSM performance on ssMNIST when using a constant image for each digit. The partially observable aspect has been removed, and LSTM successfully solves the sequence learning task. Mean accuracy and standard error shown across repeated runs.}
    \label{fig:ssmnist_fixed_acc}
\end{figure}

To confirm correctness of the LSTM baseline algorithm, we verified it is able to solve a simplified (fully observable) version of the task where the same MNIST image is used at each occurrence of a given label. Under these conditions, LSTM achieves the theoretical accuracy limit comparatively quickly, though displays volatility even after approaching this accuracy ceiling (see Figure \ref{fig:ssmnist_fixed_acc}). 
This volatility in the fixed-image regime is likely an illustration of the tendency for these sequence learning models attempting to `learn' spurious higher order transitions between sub-sequences that are not in fact predictable. 

\begin{table}
  \caption{ssMNIST results on 8x9 grammar. Accuracy is reported as mean $\pm$ one standard deviation, and max over 5 runs to account for observed inter-run variance. Theoretical ceiling on accuracy for this grammar is 88.8\%.}
  \label{tab:ssmnist}
  \begin{tabular}{lccc}
    \toprule
    Model & Params & Mean Acc & Max Acc \\
    \midrule
    LSTM (cont) & 2.6M & 80.0\% $\pm$ 9.1 & 81.4\% \\
    LSTM (mbs=100) & 2.6M & 73.4\% $\pm$ 18.2 & 82.7\% \\
    bRSM & 2.5M & 86.4\% $\pm$ 0.3 & 86.8\% \\
    bRSM (partitioned) & 1.8M & 88.8\% $\pm$ 0.1 & 88.9\% \\
  \bottomrule
\end{tabular}
\end{table}

\begin{figure}[h]
    \centering
    \includegraphics[width=\linewidth]{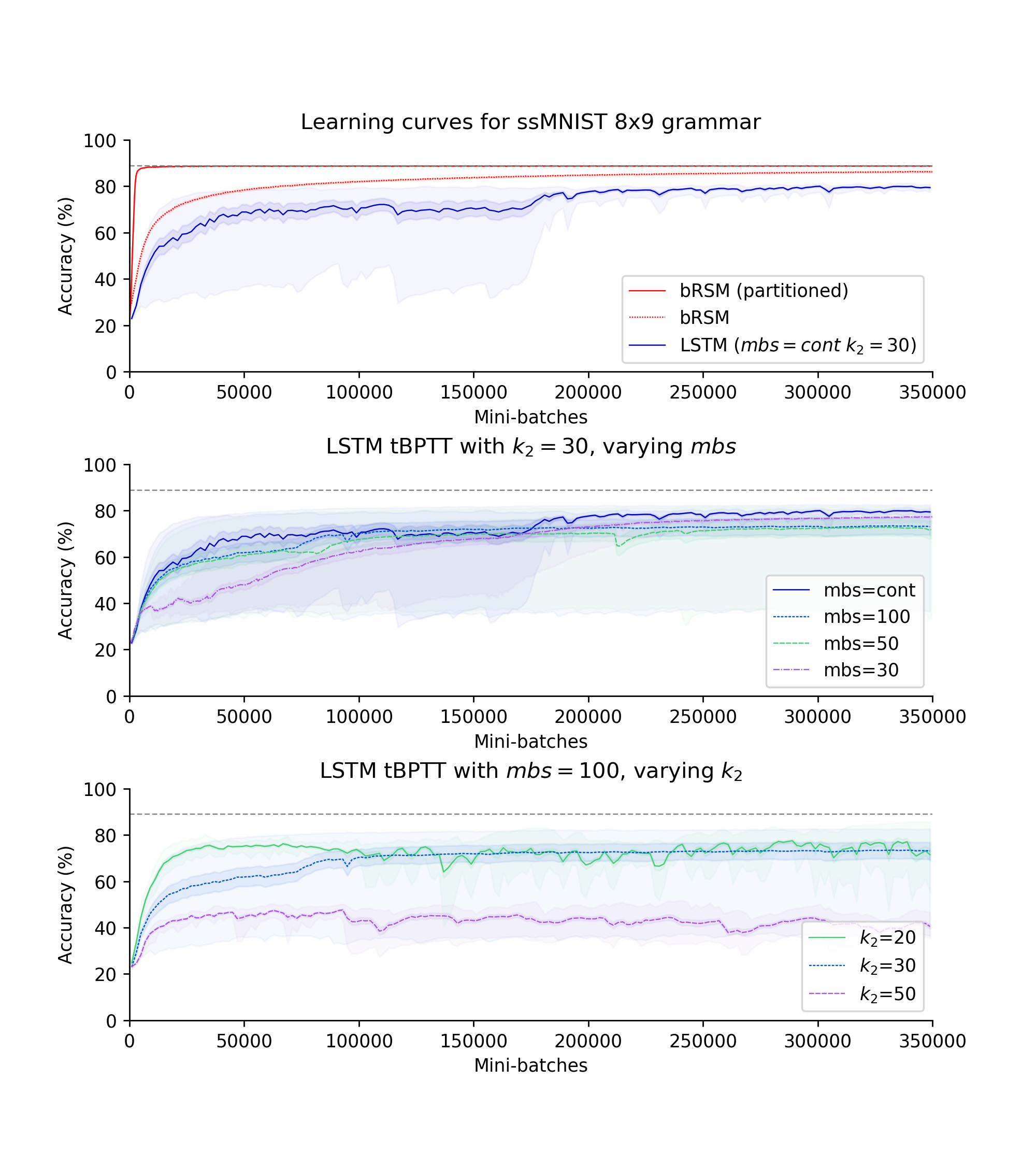}
    \caption{LSTM and bRSM performance on ssMNIST. Mean accuracy (line), standard error (shadow) and range (light shadow) across repeated runs. Gray line is theoretical accuracy ceiling for the 8x9 grammar (see Appendix \ref{appendix:sequences}).}
    \label{fig:ssmnist_acc}
\end{figure}

A second option distinct from the tBPTT parameterization was also observed to significantly impact LSTM performance. Maximum digit prediction accuracy was achieved by adjusting the training regime to periodically clear the LSTM's memory cell state. In Figure \ref{fig:ssmnist_acc}, \emph{mbs} indicates the number of time steps (and therefore mini-batches) after which we cleared the LSTM module's hidden and cell state.

Together, optimization of the backpropagation window to small finite values ($k_{2}$) and state clearing interval (\emph{mbs}) advantage the LSTM with two sources of an implicit prior on the length of salient temporal context. Intuitively, setting $k_{2}$ or $mbs$ below our grammar's sub-sequence length would make it impossible to learn high-order relationships, and too large of a value might confound the network by offering far more temporal context than is useful for learning transitions within each sub-sequence. We anticipated and confirmed that maximum accuracy would be achieved when both parameters were tuned to convey a useful prior on context while supplying a sufficient history to robustly learn the higher-order temporal relationships in the data. Results from experiments with varying configurations of tBPTT and state clearing are shown in Figure \ref{fig:ssmnist_acc} and appear to support this understanding. 

Across the variety of training regimes tested, LSTM with the continuous configuration and $k_{2}=30$ achieved the best mean accuracy across runs of 80.0\% (90.0\% of the theoretical limit for this grammar). The highest accuracy LSTM run was observed with $mbs=100$ and $k_{2}=30$, reaching 82.7\%, but inter-run variance was significantly higher in this configuration. In comparison the non-partitioned and partitioned variants of bRSM achieved 86.4\% and 88.8\% respectively, with very little inter-run variance. A summary of results is shown in Table \ref{tab:ssmnist}.

LSTM did not achieve the maximum achievable prediction accuracy even with the additional context-length clues implicitly provided by the training regime. LSTM showed slower convergence, increased volatility and lower eventual accuracy without these clues. The much better results using a constant image for each digit suggest that the combination of partial observability, sequential uncertainty and unmarked sub-sequence boundaries make this task especially difficult for conventional recurrent models. In contrast, bRSM was able to learn the partially observable sequence relationships without the need to tune hyper-parameters in accordance with the grammar's true time horizon. Furthermore, as noted by Rawlinson et al., by avoiding BPTT, RSM has an asymptotic memory use of $O(c)$, where $c$ is the number of cells in the hidden layer. This is a significant reduction from deep backpropagation models which require $O(ct)$, where $t$ is the time-horizon, even when both models have the same number of parameters. For the empirically optimal tBPTT parameterization used in this analysis $t=k_{2}=30$, which implies that 30$\times$ more memory is required. Overall, bRSM achieves better sequence learning performance than an ordinary LSTM in this partially observable condition, with less prior knowledge of the task and significantly less memory requirement.

\subsection{Language Modeling}

\subsubsection{Dataset}

Consistent with the original RSM paper, we present language modeling results using the Penn Treebank (PTB) dataset with preprocessing as per Mikolov et al. \cite{mikolov2010recurrent}. RSM's performance on this language modeling task was the weakest result of those originally reported, making it an ideal target to determine if the observed limitations could be overcome. Model evaluation was performed using the test corpus.

\subsubsection{Training Regime}

We observed that, consistent with previous findings \cite{rawlinsonLearningDistantCause2019}, the bRSM model overfits quickly to the PTB training set, as illustrated by increasing volatility and ultimately a quick rise in test loss after 40-60,000 mini-batches of training. To address this dynamic, we found it useful to pause training of the core RSM model prior to overfit, and allow the classifier network to continue training. We noted that final test set perplexity was quite sensitive to the time of pause. For the results shared here, pause epoch is considered an additional hyper-parameter. A custom stopping criteria based on the derivative of validation loss would allow for more flexible experimentation, and is planned for future work.

\subsubsection{Results}

Towards our goal of exploring the performance bounds of models under our bio-plausibility constraints, we present results from experiments with bRSM on the PTB dataset. The lowest test perplexity (103.5 PPL) was achieved using the first four additions presented in section \ref{sec:additions} (all but functional partitioning). A 7\% word cache was effective, but an ensemble of bRSM and KN5 did not significantly improve test performance. KN5 results are shown to illustrate the performance of statistically defined n-gram models.

Table \ref{tab:lm_results} reports results for the final bRSM model as well as versions of this model with each added feature ablated. bRSM, with and without the word cache, outperforms all early language modeling architectures, including ordinary (non-gated) recurrent neural language models trained with BPTT. While these results are not yet competitive with state-of-the-art deep models such as the Transformer, and modern LSTM-based approaches, they demonstrate a significant step forward for resource efficient performance.

\begin{table}
  \caption{Language modeling results. bRSM variants with each of 4 added feature ablated are shown. $\dagger$: As reported by Mikolov et al  \cite{mikolovEmpiricalEvaluationCombination}.}
  \label{tab:lm_results}
  \begin{tabular}{lccc}
    \toprule
    Model & Test PPL & No. of params \\
    \midrule
    KN5 $\dagger$ & 141.2 & -- \\
    KN5 + cache $\dagger$ & 125.7 & -- \\
    Random Forest LM $\dagger$ & 131.9 & -- \\
    RNN LM (uses tBPTT) $\dagger$ & 124.7 & -- \\
    \midrule
    LSTM & 78.9 & 13M \\
    Mogrifier LSTM & 50.1 & 24M \\
    GPT-2 & 35.7 & 1500M \\
    \midrule
    bRSM + cache & 103.5 & 2.55M \\
    $\cdot$ Non-semantic embedding & 152.6 & 2.34M  \\
    $\cdot$ Inhibition instead of boosting & 144.0 & 2.55M  \\
    $\cdot$ Non-flattened (m=800, n=3) & 112.8 & 3.36M  \\
    $\cdot$ Without cache & 112.0 & 2.55M \\
    $\cdot$ Untrained decay rate & 107.3 & 2.55M  \\
  \bottomrule
\end{tabular}
\end{table}

\subsubsection{Resource Utilization (Boosting vs Inhibition)}
\label{sec:resource_utilization}

A possible explanation for the difference in performance seen between boosting and inhibition strategies involves the strength and temporal dynamics of each. Boosting integrates a moving average of individual cell activity across hundreds of time steps, promoting the use of idle cells. In contrast, inhibition produces a strong and immediate effect where cells are fully inhibited from firing after a single activation. Both strategies aim to improve resource utilization.

One way to compare the effect of these strategies is to quantify the informational capacity of the RSM memory using layer entropy ($H_l$), which is calculated from the duty cycle as follows:

$$ H_l = \sum_{i} -d_i \log_2 d_i - (1 - d_i) \log_2 (1 - d_i) $$

We can compare layer entropy during training and at inference time with the theoretical maximum binary entropy for an RSM layer, which is a function only of layer sparseness ($s = \frac{k}{mn}$): 

$$ H_{l, max} = -s \log_2(s) - (1 - s) \log_2(1 - s) $$

\begin{figure}[h]
    \centering
    \includegraphics[width=\linewidth]{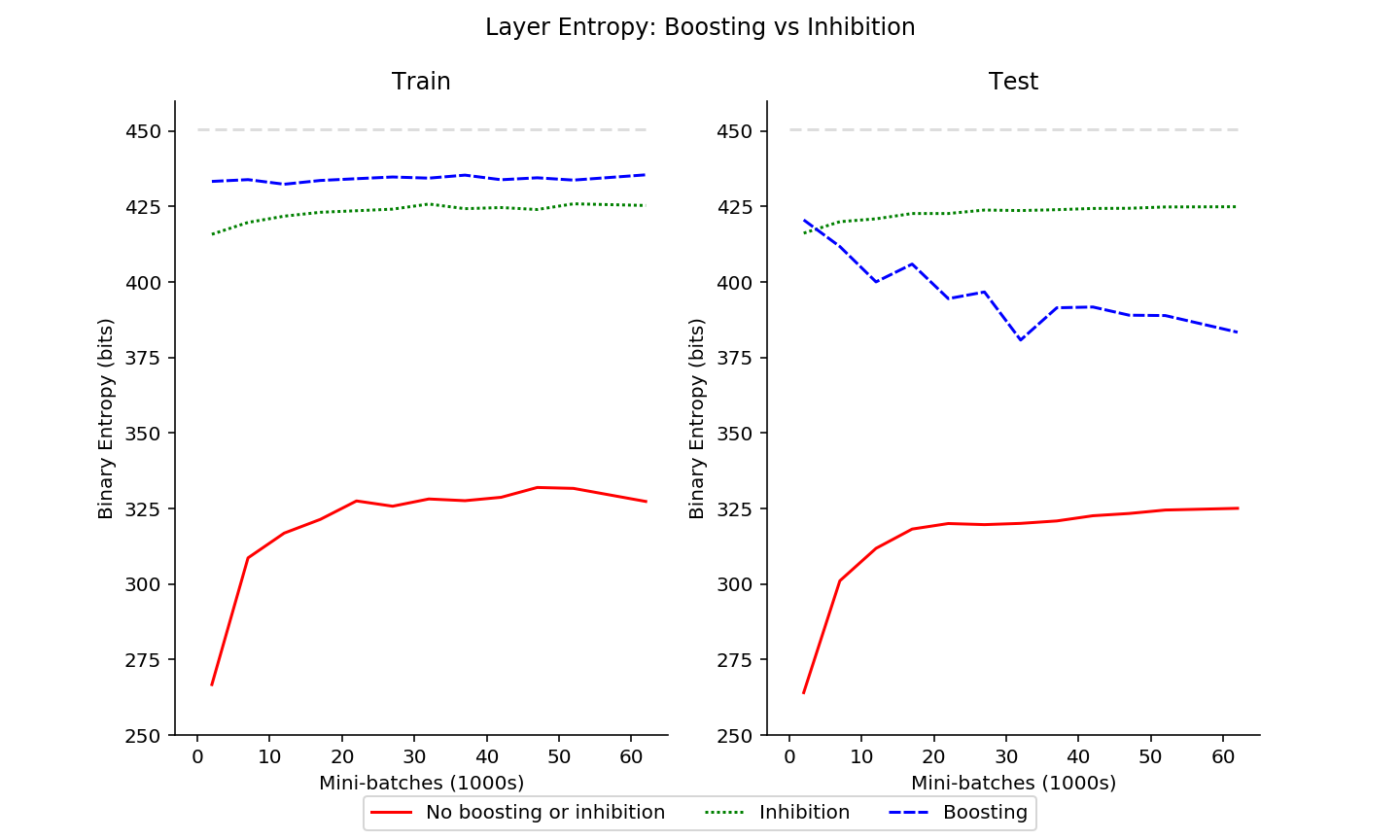}
    \caption{Layer entropy comparison of boosting vs inhibition strategy. Maximum possible layer entropy shown by dashed gray line.}
    \label{fig:entropy}
\end{figure}

In Figure \ref{fig:entropy}, we compare the time course of binary entropy for two RSM models differing only in resource utilization strategy. As expected, both strategies have the effect of increasing layer entropy compared to having no strategy to promote the use of idle cells. We note that inhibition exhibits nearly identical entropy dynamics across training and test sets---approximately 425 bits, or 93\% of maximum entropy---while the boosted model's test entropy is reduced during exposure to unseen test sequences. 

This observation supports a traditional bias-variance trade-off based understanding of the relationship between encoding entropy and generalization performance of sparse recurrent networks. In the high entropy case using inhibition, similar sequences are encoded in highly orthogonal patterns, which may support high capacity memorization. This is helpful when there is an opportunity to learn to interpret these patterns, but confounding when generalizing to unseen sequences, because similar contexts are encoded in dissimilar ways. This is consistent with our observation that inhibition produces worse perplexity and higher entropy on the test corpus.

However, some recent work has questioned the notion that high capacity function classes necessarily result in poor generalization performance \cite{belkin2018reconciling}, and so alternative explanations can be considered as well. For example, the strong inhibition of recently active cells may recruit arbitrary non-semantic encodings that struggle to generalize without implicating excessive capacity. In either case, encoding unseen sequences from the test corpus with relatively lower entropy implies that fewer unique encodings are produced. We hypothesize that the network falls back to known encodings of similar contexts, which the classifier network is able to interpret. Consequently, relatively better perplexity is observed from the lower-entropy test-corpus encoding.

\section{Conclusion}

We presented results from a sparse predictive autoencoder with a slim memory footprint, trained on a time-local error signal. As far as we're aware, this model demonstrates the best results to date on the PTB language modeling task among models not relying upon the use of memory-intensive deep backpropagation across many layers and/or time steps. Neural language models with better performance all use additional mechanisms to selectively filter and store historical state (e.g. attention and gating in Transformer and LSTM networks); our goal is not to beat them, but to show that learning rules which are local in time and space could be competitive, given further development. This work provides encouraging evidence that strong results on challenging tasks such as language modelling may be possible using less memory intensive, biologically-plausible training regimes.

We also showed that on tasks with particular characteristics---namely weak partial-observability and continual presentation of randomly-ordered sub-sequences without boundary markers---our approach outperformed the LSTM gated memory representation. This result also merits further investigation to understand the relationship between these task characteristics and local versus deep learning rules.

\bibliography{bibliography.bib}

\begin{thebibliography}{10}

\bibitem{ahmadHowCanWe}
{\sc Ahmad, S., and Scheinkman, L.}
\newblock How can we be so dense? the benefits of using highly sparse
  representations.
\newblock {\em arXiv preprint arXiv:1903.11257\/} (2019).

\bibitem{bai2018empirical}
{\sc Bai, S., Kolter, J.~Z., and Koltun, V.}
\newblock An empirical evaluation of generic convolutional and recurrent
  networks for sequence modeling.
\newblock {\em arXiv preprint arXiv:1803.01271\/} (2018).

\bibitem{belkin2018reconciling}
{\sc Belkin, M., Hsu, D., Ma, S., and Mandal, S.}
\newblock Reconciling modern machine learning and the bias-variance trade-off.
\newblock {\em arXiv preprint arXiv:1812.11118\/} (2018).

\bibitem{bojanowski2016enriching}
{\sc Bojanowski, P., Grave, E., Joulin, A., and Mikolov, T.}
\newblock Enriching word vectors with subword information.
\newblock {\em arXiv preprint arXiv:1607.04606\/} (2016).

\bibitem{gong2018frage}
{\sc Gong, C., He, D., Tan, X., Qin, T., Wang, L., and Liu, T.-Y.}
\newblock Frage: Frequency-agnostic word representation.
\newblock In {\em Advances in neural information processing systems\/} (2018),
  pp.~1334--1345.

\bibitem{haoTrainingSingleAI}
{\sc Hao, K.}
\newblock Training a single {{AI}} model can emit as much carbon as five cars
  in their lifetimes.
\newblock
  https://www.technologyreview.com/s/613630/training-a-single-ai-model-can-emit-as-much-carbon-as-five-cars-in-their-lifetimes/.

\bibitem{Hawkins2016}
{\sc Hawkins, J., and Ahmad, S.}
\newblock Why {{Neurons Have Thousands}} of {{Synapses}}, a {{Theory}} of
  {{Sequence Memory}} in {{Neocortex}}.
\newblock {\em Frontiers in Neural Circuits 10\/} (2016).

\bibitem{Hawkins-et-al-2016-Book}
{\sc Hawkins, J., Ahmad, S., Purdy, S., and Lavin, A.}
\newblock Biological and machine intelligence (bami).
\newblock Initial online release 0.4, 2016.

\bibitem{hochreiterLongShortTermMemory1997}
{\sc Hochreiter, S., and Schmidhuber, J.}
\newblock Long {{Short}}-{{Term Memory}}.
\newblock {\em Neural Computation 9}, 8 (Nov. 1997), 1735--1780.

\bibitem{makhzaniKSparseAutoencoders2013}
{\sc Makhzani, A., and Frey, B.}
\newblock K-{{Sparse Autoencoders}}.
\newblock {\em arXiv:1312.5663 [cs]\/} (Dec. 2013).

\bibitem{marcus1994penn}
{\sc Marcus, M., Kim, G., Marcinkiewicz, M.~A., MacIntyre, R., Bies, A.,
  Ferguson, M., Katz, K., and Schasberger, B.}
\newblock The penn treebank: annotating predicate argument structure.
\newblock In {\em Proceedings of the workshop on Human Language Technology\/}
  (1994), Association for Computational Linguistics, pp.~114--119.

\bibitem{melisMogrifierLSTM2019}
{\sc Melis, G., Ko{\v c}isk{\'y}, T., and Blunsom, P.}
\newblock Mogrifier {{LSTM}}.
\newblock {\em arXiv:1909.01792 [cs]\/} (Sept. 2019).

\bibitem{mikolovEmpiricalEvaluationCombination}
{\sc Mikolov, T., Deoras, A., Kombrink, S., Burget, L., and Cernocky, J.}
\newblock Empirical {{Evaluation}} and {{Combination}} of {{Advanced Language
  Modeling Techniques}}.
\newblock 4.

\bibitem{mikolov2010recurrent}
{\sc Mikolov, T., Karafi{\'a}t, M., Burget, L., {\v{C}}ernock{\`y}, J., and
  Khudanpur, S.}
\newblock Recurrent neural network based language model.
\newblock In {\em Eleventh annual conference of the international speech
  communication association\/} (2010).

\bibitem{mikolovExtensionsRecurrentNeural2011}
{\sc Mikolov, T., Kombrink, S., Burget, L., {\v C}ernock{\'y}, J., and
  Khudanpur, S.}
\newblock Extensions of recurrent neural network language model.
\newblock In {\em 2011 {{IEEE International Conference}} on {{Acoustics}},
  {{Speech}} and {{Signal Processing}} ({{ICASSP}})\/} (May 2011),
  pp.~5528--5531.

\bibitem{rawlinsonLearningDistantCause2019}
{\sc Rawlinson, D., Ahmed, A., and Kowadlo, G.}
\newblock Learning distant cause and effect using only local and immediate
  credit assignment.
\newblock {\em arXiv:1905.11589 [cs, stat]\/} (May 2019).

\bibitem{oordWaveNetGenerativeModel2016}
{\sc van~den Oord, A., Dieleman, S., Zen, H., Simonyan, K., Vinyals, O.,
  Graves, A., Kalchbrenner, N., Senior, A., and Kavukcuoglu, K.}
\newblock {{WaveNet}}: {{A Generative Model}} for {{Raw Audio}}.
\newblock {\em arXiv:1609.03499 [cs]\/} (Sept. 2016).

\bibitem{vaswaniAttentionAllYou2017}
{\sc Vaswani, A., Shazeer, N., Parmar, N., Uszkoreit, J., Jones, L., Gomez,
  A.~N., Kaiser, {\L}., and Polosukhin, I.}
\newblock Attention is all you need.
\newblock In {\em Advances in neural information processing systems\/} (2017),
  pp.~5998--6008.

\bibitem{williamsEfficientGradientBasedAlgorithm}
{\sc Williams, R.~J., and Peng, J.}
\newblock An efficient gradient-based algorithm for on-line training of
  recurrent network trajectories.
\newblock {\em Neural computation 2}, 4 (1990), 490--501.

\end{thebibliography}
\bibliographystyle{acm}

\appendix

\section{ssMNIST Sequences} 
\label{appendix:sequences}

\subsection{``8x9'' sequence generation}

\subsubsection{Sub-sequences}

The ``8x9'' grammar used in reported results is composed of the following sub-sequences, shown in rows below:

\vspace{7pt}

\noindent
2, 4, 0, 7, 8, 1, 6, 1, 8 \\
2, 7, 4, 9, 5, 9, 3, 1, 0 \\
5, 7, 3, 4, 1, 3, 1, 6, 4 \\
1, 3, 7, 5, 2, 5, 5, 3, 4 \\
2, 9, 1, 9, 2, 8, 3, 2, 7 \\
1, 2, 6, 4, 8, 3, 5, 0, 3 \\
3, 8, 0, 5, 6, 4, 1, 3, 9 \\ 
4, 7, 5, 3, 7, 6, 7, 2, 4 \\

Note that several two and three-digit transitions are shared between sub-sequences, but no two sub-sequences share the same first two digits.

\subsubsection{``8x9'' grammar accuracy ceiling calculation}

Given the semi-deterministic nature of the sample generating process and grammar defined, we can calculate the theoretical limit on prediction accuracy as follows.

\vspace{5pt}

\noindent
\emph{\nth{1} digit}: predict 2 at P=3/8.\\
\emph{\nth{2} digit}:

\indent Following 2: predict $\{4, 7, 9\}$ uniformly \\
\indent Following 1: predict $\{2, 3\}$ uniformly \\
\indent All remaining deterministic \\
\indent $P = (3/8 * 1/3) + (2/8 * 1/2) + (3/8 * 1)$ \\
    
\noindent
\emph{Remaining digits}: deterministic conditioned on first 2 digits.

\noindent
Correct predictions per sequence: $(3/8 + [(3/8 * 1/3) + (2/8 * 1/2) + (3/8 * 1)] + 7) = 8$\\
Accuracy ceiling: $8 / 9 = \textbf{88.88\%}$

\section{Model Details}
\label{appendix:model_details}

\subsection{Description of hyper-parameters}

\textbf{Probability of forgetting} is a parameter used to expose the network to novel sequences by clearing the memory state at randomized intervals. This is parameterized by $\mu$, the probability at each time step, and for each training sequence, of clearing the hidden state.

\textbf{Boost strength} controls the influence of the per-cell boost computation within the top-k algorithm. It is a non-negative parameter, and disables boosting when set to 0.

\textbf{Boost strength factor} allows an exponential decay of boost strength, which has been show to stabilize training.

\textbf{Uniform mass weight} controls the interpolation of a uniform distribution with the output of the main model. The final distribution used to compute loss is calculated as a weighted average of each interpolated model distribution.

\textbf{Word cache weight} controls the interpolation of the simple word cache used in some experiments. 

\textbf{Word cache decay rate} controls the decay of the word cache, which is implemented as a tensor with dimension equivalent to the size of the corpus vocabulary. After each token is observed, its index in the cache is set to 1. The cache is decayed according to this parameter on each step.

\subsection{Hyper-parameters used}

Tables \ref{tab:hparam_lm} and \ref{tab:hparam} list the configurations for hyper-parameters for the language modeling and ssMNIST experiments respectively.

\begin{table}[hbt!]
  \caption{Hyper-parameters used (language modeling)}
  \label{tab:hparam_lm}
  \begin{tabular}{lcc}
    \toprule
    Description & Symbol & Value \\
    \midrule
    Batch size & -- & 300 \\
    Probability of forgetting & $\mu$ & 0.025 \\
    Decoder L2 regularization & --  & 0.00001  \\
    No. of groups / mini-columns & $m$ & 1500 \\
    No. of cells per group & $n$ & 1 \\
    Number of winning groups / cells & $k$ & 80 \\
    Boost strength & $\beta$ & 1.2 \\
    Boost strength factor & -- & 0.85 \\
    Predictor hidden size & -- & 1200 \\
    Uniform mass weight & -- & 0.01 \\
    Word cache weight & -- & 0.07 \\
    Word cache decay rate & -- & 0.99 \\
  \bottomrule
\end{tabular}
\end{table}

\begin{table}[hbt!]
  \caption{Hyper-parameters used (ssMNIST)}
  \label{tab:hparam}
  \begin{tabular}{lcc}
    \toprule
    Description & Symbol & Value \\
    \midrule
    Batch size & -- & 300 \\
    Decoder L2 regularization & --  & 0.0  \\
    No. of groups / mini-columns & $m$ & 1000 \\
    No. of cells per group & $n$ & 1 \\
    Number of winning groups / cells & $k$ & 120 \\
    Boost strength & $\beta$ & 1.2 \\
    Boost strength factor & -- & 0.85 \\
    Predictor hidden size & -- & 1200 \\
  \bottomrule
\end{tabular}
\end{table}

\section{Word Embeddings}

\subsection{Synthetic Embedding}
\label{appendix:synthetic_embedding}

The synthetic embedding was constructed as per the original RSM work as follows:

For each $i^{th}$ word in the corpus, a 28-dimensional binary embedding is generated. The binary vector is constructed as the 14-bit left-filled binary encoding of the vocabulary index $i$, concatenated with its inverse.

For example, the second word in the corpus, $vocab[1]$, would be embedded as $0000000000000111111111111110$, and the \nth{100} words in the corpus, $vocab[99]$, would be embedded as\\ $0000000110001111111110011100$.

\subsection{FastText Embedding}
\label{appendix:fasttext}

We used FastText's unsupervised training method \footnote{Code for generating FastText embeddings on custom corpora is available at \url{https://github.com/facebookresearch/fastText}} to generate a single fixed embedding vector for each word in the PTB vocabulary. We used the skipgram model with learning rate ($lr$) of 0.1, a vector dimension ($dim$) of 100, minimal number of word occurrences ($minCount$) of 1, softmax loss ($loss$), and trained for 5 epochs ($epoch$). Embeddings were stored in a static dictionary once generated and treated as inputs to the RSM network.

\end{document}